\documentclass[conference]{IEEEtran}
\IEEEoverridecommandlockouts
\usepackage{cite}
\usepackage{amsmath,amssymb,amsfonts}
\usepackage{algorithmic}
\usepackage{graphicx}
\usepackage{textcomp}
\usepackage{xcolor}
\usepackage{caption}
\usepackage{subcaption}
\usepackage{multirow}
\usepackage{makecell}

\def\BibTeX{{\rm B\kern-.05em{\sc i\kern-.025em b}\kern-.08em
    T\kern-.1667em\lower.7ex\hbox{E}\kern-.125emX}}
\begin{document}

\title{Deep Latent Defence\\
{}
\thanks{This work was funded by an Industrial CASE studentship jointly between the UK Engineering and Physical Science Research Council (EPSRC) and Airbus.}
}
\author{\IEEEauthorblockN{Giulio Zizzo\IEEEauthorrefmark{1}\IEEEauthorrefmark{2}, Chris Hankin\IEEEauthorrefmark{1}\IEEEauthorrefmark{2}, Sergio Maffeis\IEEEauthorrefmark{2} and Kevin Jones\IEEEauthorrefmark{3}}
	\IEEEauthorblockA{\IEEEauthorrefmark{1}Institute for Security Science and Technology (ISST), Imperial College London}
	\IEEEauthorblockA{\IEEEauthorrefmark{2}Department of Computing, Imperial College London}
	\IEEEauthorblockA{\IEEEauthorrefmark{3}Airbus}
	\IEEEauthorblockA{g.zizzo17@imperial.ac.uk\qquad c.hankin@imperial.ac.uk\qquad sergio.maffeis@imperial.ac.uk\qquad kevin.jones@airbus.com}
	}

\maketitle

\begin{abstract}
Deep learning methods have shown state of the art performance in a range of tasks from computer vision to natural language processing. However, it is well known that such systems are vulnerable to attackers who craft inputs in order to cause misclassification. The level of perturbation an attacker needs to introduce in order to cause such a misclassification can be extremely small, and often imperceptible. This is of significant security concern, particularly where misclassification can cause harm to humans. 

We thus propose Deep Latent Defence, an architecture which seeks to combine adversarial training with a detection system. At its core Deep Latent Defence has a adversarially trained neural network. A series of encoders take the intermediate layer representation of data as it passes though the network and project it to a latent space which we use for detecting adversarial samples via a $k$-nn classifier. We present results using both grey and white box attackers, as well as an adaptive $L_{\infty}$ bounded attack which was constructed specifically to try and evade our defence. We find that even under the strongest attacker model that we have investigated our defence is able to offer significant defensive benefits. 
\end{abstract}


\section{Introduction}

With deep learning systems showing impressive results, and their continued adaptation in areas from autonomous vehicles \cite{Semantic_Segmentation}, disease detection \cite{Medical_Survey}, and cyber security \cite{LSTM_Networks} their security and integrity is of increasing importance. 

Unfortunately, deep learning methods have shown to be brittle against attack, and even with imperceptible perturbations a machine learning system can cause misclassification with a high degree of confidence\cite{Adverserial_Intriguing}. This is now extending beyond images, with adversarial samples being created for audio \cite{Adversarial_Audio}, cyber\cite{Adverserial_Feng}\cite{Adversarial_Malware}, and reinforcement learning \cite{Adversarial_RL} domains. 

Therefore there is a need for machine learning capabilities to provide reliable and secure predictions in an adversarial environment. There have been many approaches investigated in the literature, from adversarial training \cite{Madry}, Bayesian detection mechanisms \cite{Adversarial_BayesianEyes}, input processing \cite{Feature_Squeezing}, and provable guarantees \cite{AI2}. However, so far there has not been a defensive mechanism that offers scalable and robust defences to all possible attacks.

In this paper we propose a method by which we combine adversarial training and a new detection mechanism. Our defensive mechanism projects the intermediate layer output from a neural network to a lower dimensional latent space created by an encoder. By crafting the latent space to cluster the classes according to the $L_2$ distance we can apply a $k$-nn algorithm to compare the training data embeddings to test time samples. Test time samples which are surrounded by training data of a different class are regarded as suspicious and declared adversarial. When combined with adversarial training this proves challenging for an adversary to optimise over and we show high levels of robustness.   
 
The contributions of this paper are as follows:
\begin{itemize}
	\item We propose a defensive method which effectively combines adversarial training with detection based on prediction credibility utilising intermediate layer information. 
	\item We evaluate the defence against a range of attackers of different strengths, including one specifically tailored towards circumventing our defence. 
\end{itemize}
\section{Background}

\subsection{Neural Networks}

A neural network is a function, $F$, which uses learned parameters $\theta$ to map input $x\in \mathbb{R}^n$ to output $y \in \mathbb{R}^m$. The output $y$ in the case of classification is a probability distribution over $m$ classes, and the highest probability is taken as the predicted class of input $x$, i.e $C(x) = \text{arg max} F(x)$. In classification the cross entropy loss is frequently used as a minimisation objective to guide the optimisation of the parameters. During the testing phase the neural network makes predictions on data presented to it. It is at this stage that an attacker can construct evasion attacks which are consistently misclassified despite the data retaining the original semantic significance.   

\subsection{Adversarial Examples}

A attacker goal when attacking a classification network is when given a legitimate test example $x$, to construct an input $x^*$ such that $C(x^*) \neq C(x)$. The adversarial sample $x^*$ is defined to be ``close" to the test sample it was constructed from according to some distance metric $d(x,x^*)$. The case of $C(x^*) \neq C(x)$ would represent an untargeted attack while a targeted attack aims to make $C(x^*) = t$ when $t \neq C(x)$. 

To define a complete attacker one needs to specify 1) the knowledge the attacker has regarding the system, 2) the level of perturbation allowed, and 3) the algorithm employed in crafting the adversarial samples. We shall go into each in more detail below.

\subsection{Attacker Knowledge}

The attacker knowledge places constraints on the available techniques that can be used for adversarial sample crafting. On one end, white box attackers have access to all of the model parameters, thresholds, random seeds, as well as the training and testing datasets. On the other hand, black box attackers do not know the model architecture or parameters. Between these two extremes lie grey box models which are most commonly defined as an attacker which is not aware of any defensive mechanism in place \cite{Magnet}\cite{Adversarial_Shield}.

\subsection{Perturbation Levels}

The attacker is usually limited in the amount of perturbation that can be added to an input. This is generally defined to be limited by some $L_p$ norm. Commonly used norms are the $L_{\infty}$, $L_{0}$, or $L_{2}$. The $L_{\infty}$ norm determines the maximum change that can be applied to any feature. The $L_{0}$ norm specifies the total number of features that can be altered. Finally the $L_{2}$ norm represents the euclidean distance between the input that the attacker starts with and the adversarially crafted sample. 

In the case of the image domain, there is an acknowledged deficit in these norms to describe human perception\cite{TheRulesOfTheGame}. Adversarial attacks can be crafted with translation and rotation\cite{Adversarial_Roation} or patches\cite{Adversarial_Patch} which would have extremely large $L_p$ norms and yet be indistinguishable when viewed by a human. Obtaining more meaningful and security motivated constraints is an active area of research and we consider it beyond the scope of this work.

\subsection{Crafting Algorithms}

There is a wide range of attack algorithms which have been developed. Here we will go over the methods which are available to an attacker which has access to the model parameters and can thus obtain gradient information although weaker gradient free attacks are possible\cite{Adversarial_Genattack}.

\subsubsection{One Step Methods}

One step adversarial attacks involve crafting an input to maximise a loss function in a single step proportional to the sign of the gradient \cite{Adversarial_Explain}. This method is named the Fast Gradient Sign Method (FGSM) and is defined as, 
\begin{equation}
\label{FGS}
x^* = x + \epsilon \textnormal{sign}(\nabla_x J(\theta,x,y))
\end{equation}

where $J$ is the neural network's loss function. The parameter $\epsilon$ determines the level of perturbation added to the input. Despite its speed and simplicity, against and undefended network it can cause severe loss in accuracy.

\subsubsection{Iterative Methods}

Iterative methods repeatedly apply the gradient sign with a step size determined by $\alpha$, 
\begin{equation}
\label{Iterative}
x_{t+1} = x_t + \alpha\textnormal{sign}(\nabla_x J(\theta,x,y)).
\end{equation}

This type of attack is much stronger then the single step method and networks which have been defended against the FGSM can be vulnerable to iterative attacks. 

\subsubsection{Optimization Methods}

Optimisation methods directly optimise the distance between the real and adversarial example as well as target misclassification. One of the more advanced methods as proposed by \cite{CW} solves:
\begin{equation}
\underset{x^*}{\text{arg min }} ||x^* -x||_p -c f(x^*,y),
\end{equation}

where $f$ is a function chosen so that $f(x^*,y) \leq 0$ only if the target network misclassifies $x^*$ and $p$ is the chosen norm. The parameter $c$ acts as a weighting term. This formulation will generally cause the misclassified sample to have a lower distortion.

\section{Defences}

There are two broad strategies being researched for combating adversarial samples. The first, aims to directly make the underlying neural network inherently robust, while the second seeks to detect adversarial samples.

\subsection{Adversarial Training}

Adversarial training was introduced in \cite{Adversarial_Explain} and uses the FGSM to augment the network's loss function to 

\begin{equation}
\begin{split}
\tilde{J}(\theta,x,y) = \alpha J(\theta,x,y)  + (1-\alpha)J(\theta,x^*,y)
\end{split}
\end{equation}

where $\alpha$ is a hyperparameter set to 0.5. This defence trains the network on both normal and adversarial examples however it can be broken if the attacker uses a more sophisticated attack even when operating under the same $L_{\infty}$ norm. A very robust evolution of the defence was proposed in \cite{Madry} where attacks are crafted using an iterative method and a neural network trained on them. 

The weakness of adversarial training methods is that they strongly prescribe the type of attacker to which they are robust to. If an attacker introduces a different distortion metric, or a slightly larger perturbation than expected, then the level of defence offered rapidly drops.

\subsection{Detection by Uncertainty}

A class of detection methods use uncertainty estimates. Neural networks do not by default give reliable uncertainties associated with a given prediction. There is a belief that the final softmax outputs of a neural network represent model confidence, however the softmax outputs are poorly calibrated and often give misleading interpretations.

There have been many approaches recently in building Bayesian neural networks to obtain principled uncertainty estimates. One of the most widely used approaches to tackling this problem was proposed by \cite{Gal_dropout} and used dropout at test time and examined the variation in softmax outputs as a uncertainty metric. Drawbacks with such an approach are that it is known to underestimate the uncertainty \cite{Understanding_Uncertinty}. However, Bayesian methods are showing promising results in certain areas. For example although adversarial samples that evaded the defence in \cite{Adversarial_Detecting_From_Artifacts} can be created, they required a large level of distortion \cite{Adversarial_NotEasilyDetected}.

Most related to our work is \cite{dkNN} which introduced the idea of applying a distance measure between the training and test data throughout the neural network's intermediate layers. Should test time data deviate significantly from the training data distribution then it is marked as low credibility and declared adversarial. We extend and modify these principles to provide a high level of defence against a range of adversarial attacks, by explicitly optimising the intermediate layers to allow for improved searching in our chosen distance metric.

\section{Deep Latent Defence}

We now seek to investigate whether adversarial training and prediction uncertainty can complement each other. Frequently, detection based approaches can achieve extremely high detection accuracies. However, they are usually found to be flawed in some respect, as several detection schemes are then broken or fooled by modifications of the attack that they were evaluated against. Conversely, adversarial training is robust but does not usually offer protection as high as is initially shown to be possible with detection methods \cite{Adversarial_NotEasilyDetected} and adversarial training is more limited in the range of attacks it can protect against. 

\begin{figure}[t]
	\centering
	\includegraphics[trim = 0mm 145mm 80mm 0mm, clip, scale=0.33]{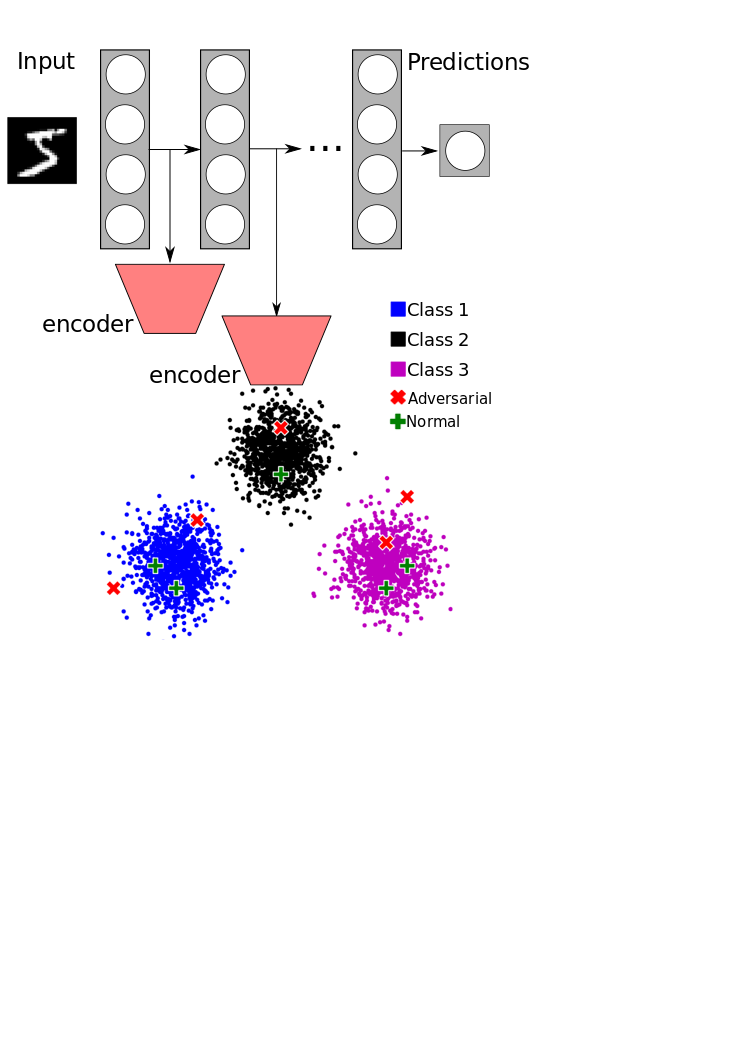}
	\caption{Overview of Deep Latent Defence during deployment illustrating an example case with 3 potential classes in 2D space for a second layer encoder. The classifier network takes an input and predicts a particular class. Seeing the classes of nearby training points in embedding space  then we can measure how much the training data supports the network's prediction for the given input. Green plus signs illustrate test data which was predicted the same class as it's $k$-nn training points and is therefore normal test data. Red crosses illustrate discrepancies between training data and predicted class of the sample and so is declared adversarial.}
	\label{Defence_Overview}		
\end{figure}

Our defensive system, which we name Deep Latent Defence, has at its core a neural network to be defended. Around this neural network a series of defensive encoders are placed which take in intermediate layer output as a test sample passes though the network and projects it into a lower dimensional latent space. We pre-compute the training data embeddings in each encoder's latent space prior to deployment. By comparing the network's prediction of the test sample with the classes of its $k$-nn training points in each encoder's latent space we can gauge how much the training data is in support of a prediction. Should the network predict a specific class $t$ and yet the test sample's embedding is surrounded by training data points of class $c$, with $c \neq t$, then the input is flagged as suspicious. An overview of the architecture is shown in Fig. \ref{Defence_Overview}.

We thus follow two lines of investigation: firstly, examining a way of combining adversarial training with our proposed detection mechanism. It is known that adversarial training leads to more robust feature selection \cite{No_Free_Lunch}, and thus should be able to supply more meaningful information to a uncertainty based detector making it more secure. Secondly, we investigate what additional information can be gained from the internal layers of a neural network rather than just examining the input and the final prediction.

\subsection{Training Phase}
\label{sec:Training}
We begin by adversarially training a convolutional neural network to conduct image classification. We use the projected gradient descent method (PGD) proposed by \cite{Madry} to generate adversarial samples. The PGD method as conducted in \cite{Madry} iteratively constructs adversarial examples following equation \ref{Iterative}, however the initial starting sample $x_0$ has Gaussian noise added to it to give different starting locations for the PGD method. This makes the networks less prone to over-fitting and more robust.       

On MNIST this baseline network achieves 98.7\% accuracy on test samples and 91.31\% accuracy on adversarial samples crafted with the PGD method using 40 iterations, a step size of 0.001, and a maximum perturbation of 0.3. Our baseline performance for SVHN was 94.76\% when trained with a maximum perturbation of 0.05, a step size of 0.005 and 20 iterations.

To improve on this we turn to our ideas of uncertainty and internal layer information. For a given layer $N$ its output, $O_N$ is fed into an encoder-decoder model, $A_N$ with parameters $\theta_N$ which returns its decoded representation back into the classifier neural network $F$. This encoder-decoder is trained with the objective of maximum classification accuracy on the convolutional neural network with cross entropy loss $J_{CE}$, 

\begin{equation}
\underset{\theta_N}{\text{arg min }} J_{CE}(F(A_N(O_N,\theta_N)), y).
\end{equation}

where $y$ are the class labels for the data for the data.

Note that $F$'s parameters, which have been previously trained, remain fixed during this second training phase. This architecture yields a latent space that is much more suited to $k$-nn clustering compared to simply reconstructing an input. We originally experimented with also including a reconstruction of the original input in the training objective, however it ultimately offered little additional benefit to the overall detection mechanism as a white box attacker can easily optimise over it. Thus, to speed up training and test time performance the reconstruction objective was dropped. This is the principal motivation from refraining to use the term `autoencoder' to refer to the encoder-decoder model, as autoencoders usually incorporate a reconstruction term in the loss function.  

To now introduce notions of uncertainty and build a detection system, we use a modified version of Deep $k$-nn \cite{dkNN} more explicitly tuned for adversarial sample detection applied in the latent space created by the encoder-decoder. One advantage of this is that we can explicitly craft the latent space for $k$-nn detection in $L_2$ distance. This differs from \cite{dkNN} which directly used the intermediate layers in the classification network for both $k$-nn classification using hash collisions and neural network classification. While a normal latent space (or intermediate layer output) is not inherently suited for a $k$-nn using an $L_2$ metric being applied directly, to encourage well clustered classes in latent space with respect to the $L_2$ distance we augment the training objective with a contrastive loss function \cite{Constrastive_Loss}. 

To use the contrastive loss function we create pairs of datapoints with each pair containing two data samples of the same class, or two datapoints belonging to different classes.  They receive a label $Y \in [0,1]$ which flags if both items in the pair are of the same class. We create equal numbers of same class pairs ($Y=0$) and different class pairs ($Y=1$). To train the defensive encoder on layer $N$, both data samples in a pair are passed though the classifier network's layers 1 through $N$. At this point the pair is split with one data sample in the pair going through the encoder and the other through an auxiliary encoder which provide embeddings $X_1$ and $X_2$ respectively. The auxiliary encoder has the same architecture as the encoder in this implementation, but is only involved in the task of minimising the contrastive loss. As the tasks of the encoder and auxiliary encoder are slightly different, and to allow greater model flexibility we do not use weight tying between the two encoders. Additionally we are not interested in certain benefits which come from weight tying, such as symmetric prediction when the datapoints are swapped in a pair, as we do not use $D(X_1,X_2)$ for detection.

The contrastive loss function, $J_c(Y,X_1,X_2)$, is therefore defined as,   

\begin{equation}
\begin{split}
J_c(Y,X_1,X_2) = (1-Y)\frac{1}{2}D(X_1,X_2)^2 + \\
Y\frac{1}{2}{max(0,m - D(X_1,X_2))}^2 
\end{split}
\end{equation}

where $D(X_1,X_2)$ is the $L_2$ distance between the embeddings and $m$ is a margin by which we want the classes to be separated; set to 1 in our experiments. This gives the overall loss function, $J_{T}$, to be: 

\begin{equation}
 J_{T} = J_{CE} + J_c.
\end{equation}

\begin{figure}[t]
	\centering
	
	\includegraphics[trim = 0mm 247mm 0mm 0mm, clip, scale=0.33]{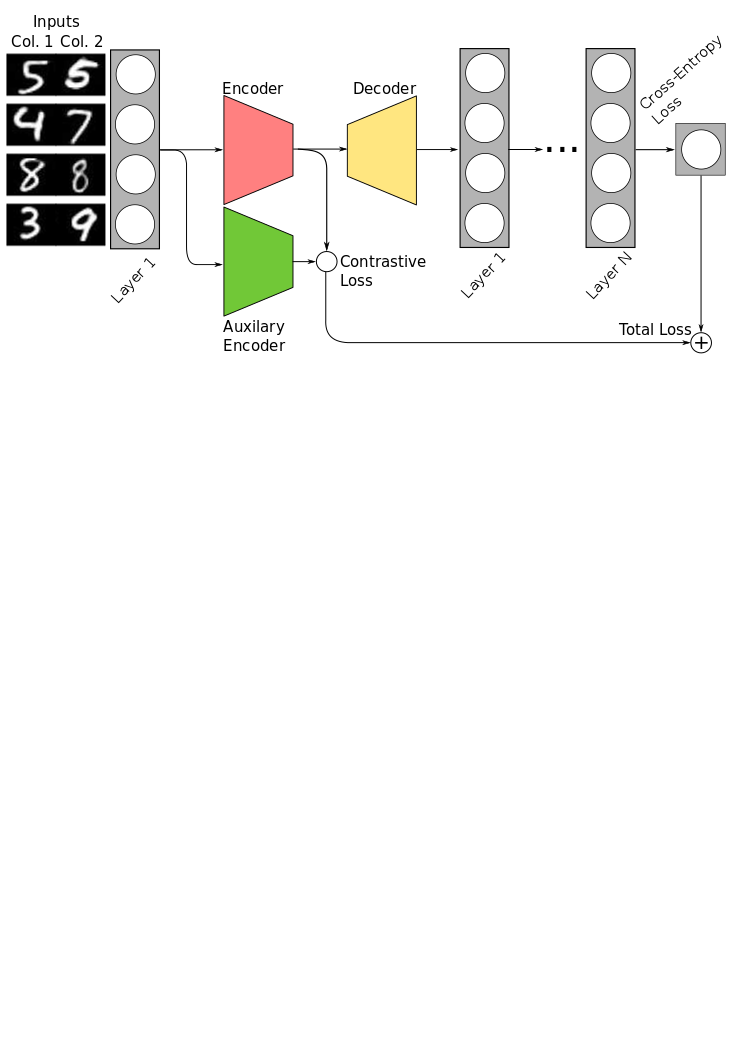}
	\caption{Training set up for our defensive system. Here we show how the encoder for layer 1 is trained. We create pairs of data which are made up in equal numbers of same and different classes. A given pair goes through the first layer of our classifier network. The pair is now split with the first item in the pair (column 1 in the diagram) going through the encoder, while the second item (column 2) going through the auxiliary encoder. Both the auxiliary and normal encoder's outputs are then used to compute the contrastive loss. However, only the encoder's output is then passed through the decoder which feeds its output back into the classifier network from layer 1. To train a defensive encoder for network layer $N$ the diagram would therefore change to have layers 1 to $N$ on the left of the encoder, auxiliary encoder, and decoder. The whole of the network $F$ will be to the right of the decoder.}
	\label{Train_Time_Overview}		
\end{figure}

Similarly to our base classification network the auxiliary encoder and the encoder-decoder pair are adversarially trained. After training is completed we discard the auxiliary encoder and the decoder to have a final architecture as shown in Fig. \ref{Defence_Overview}. This training procedure is shown for a first layer set-up in Fig \ref{Train_Time_Overview}.

\subsection{Detection Phase}

To detect adversarial samples we first project the training data into the latent spaces created by our encoders. Then on a given test point we obtain the network's predicted class and the embedding of the test sample in latent space. We compute the closest 10 $k$-nn training points in latent space. We then use the idea of non-conformity introduced in \cite{dkNN}, which is defined as the number of nearby training points with a label $i$ different to the label $j$ of the test point that is assigned by the classification network,
\begin{equation}
\beta_N(x,j) = |i \in \Omega_{N} : i \neq j|.
\end{equation}

where $\Omega_{N}$ is the multi set of labels for the $k$-nn training points in embedding space created by encoder $N$.  Each encoder hence generates a non-conformity score $\beta_N$.

\begin{table*}[t]
	\begin{center}
	\captionsetup{justification=centering}
		\caption{Defence performance of our algorithm. Base Accuracy corresponds to an adversarially trained neural network as described in Section \ref{sec:Training}. The difference between the Base Accuracy and Robustness corresponds to the contribution of the detector.}
		\begin{tabular}{| c | c | c |c |c| c| }
			\hline
			Dataset & Attack & Attack Parameters & Base Accuracy & Robustness & ROC AUC  \\ 
			\hline \hline
			\multirow{4}{*}{MNIST} & FGSM & $\epsilon = 0.3$ &  95.64\% &  98.40\%&  0.994   \\ 
			                       & PGD &  $\epsilon = 0.3, \alpha = 0.01, i =100$ & 91.20\% & 97.56\%& 0.99  \\  
			                       & CW & $i = 2000$ & 11.73\% & 96.10\% & 0.99  \\
			                       & Adaptive & $\epsilon = 0.3$ & 92.65\% & 95.62\% & 0.982  \\
			\hline \hline

			\multirow{4}{*}{SVHN} & FGSM & $\epsilon =  0.05$ & 65.56\%  &94.59\% & 0.971   \\ 
			                       & PGD &  $\epsilon = 0.05, \alpha = 0.005, i =100$ & 22.59\% & 79.69\%&  0.906  \\  
			                       & CW & $i = 1000$ & 3.24\% &99.41\% &0.993  \\
			                       & Adaptive & $\epsilon =  0.05$ & 28.89\% &51.32\% &0.811  \\
			\hline 
			
		\end{tabular}
		\label{GreyBox}
	\end{center}
\end{table*}

We now perform an additional modification to the work from \cite{dkNN} in order to more actively take steps to reduce false positives by pruning the training data included in latent space. Once training is complete we predict classes for the training data by passing it through the classification network with a intermediate layer encoder-decoder placed in layer $N=1,...,M$. Fig \ref{Train_Time_Overview} shows the architecture for $N =1$ with a first layer encoder-decoder. Training data that has its label incorrectly predicted is not included in the $N$\,th layer encoder's latent space. This filtering has a beneficial effect on both true and false positive rates. We further experimented with including adversarial samples that have been correctly classified by the network, however it resulted in a detrimental effect to performance. 

During test time, data seen by the network will be assigned a non-conformity value. A threshold for acceptable levels of non-conformity is needed for us to declare a sample as adversarial or normal. To determine this non-conformity threshold we split our test data. A calibration set of size 750 is removed and not evaluated on, and it is used to compute a non-conformity threshold for each encoder. The non-conformity value is used as a measure of how credible a test sample's predicted class is. A highly unusual sample will have a large non-conformity value and should it be greater then our threshold it is flagged as adversarial.   

\section{Experimental Results}

We present our experimental results which have been tested on MNIST and SVHN. For our classifier networks we used a 4 layer convolution network for MNIST and an 8 layer convolution network for SVHN. We used three encoders on the first three layers of our MNIST classifier and used two on the first two layers in the SVHN case. The aggregate non-conformity score resulting from the encoders is taken and if it exceeds a given threshold a test point is declared anomalous. We tried different configurations of the encoders and found this to give us the best results, as encoders placed deeper into the network have a diminishing effect because the deeper a adversarial sample travels, then the more like the target class it will resemble.

In determining the latent space size we ran a grid search and, as expected, found that the larger the latent space the fewer false positives occur although less adversarial samples are detected. This matches general intuition as a larger latent space will retain more information from the source, preserving both class information and adversarial perturbation. In our experiments we used a latent dimension of size 10 for MNIST and 100 for SVHN.
  
For our results we say the system is robust against an adversarial sample if:
\begin{enumerate}
	\item The attack is successfully classified.
	\item The attack is not successfully classified, but is detected by our detection system.
\end{enumerate}

This metric of either successful or detection is referred to as \textit{robustness}. In the case where an adversarial sample is misclassified by the network and yet is detected then we say our detection system gave a true positive result.  

A false positive is defined as a test set sample with no malicious perturbations added and yet is flagged as adversarial. We consider that the false positive rate should be measured on normal test time data encountered by the network during regular deployment. Therefore, we do not regard an adversarial sample which has been correctly classified, and flagged by our detection system to be a false positive. We would argue that an end user, particularly in security sensitive domains, would always want to be alerted to an adversarial attack, even if it is correctly classified by the network.

\subsection{Grey Box Attacker}

A grey box attacker represents an attacker who knows the target model's architecture and parameters, but otherwise unaware of defensive mechanisms in place. We run different $L_{\infty}$ bounded attacks as commonly seen in literature as well as the Carlini Wagner (CW) $L_2$ \cite{CW} attack on our network. As the CW attack uses a different norm, it circumvents the adversarial training employed. However, such attacks were highly visible in the latent space for our experiments. For the PGD method we report the worse-case performance of our defensive system using 5 random restarts. The results for the grey box attacks are shown in Table \ref{GreyBox}. In addition to the robustness metrics conditioned on a specific detection threshold we show the receiver operating characteristic (ROC) area under curve (AUC) scores as a metric for our true positive (TP) and false positive (FP) rates at varying detection thresholds.

We can see from Table \ref{GreyBox} that the balance between the latent space detection system and the underlying neural network accuracy in contributing to the overall robustness shifts as the attack becomes more powerful or changes from a $L_{\infty}$ to a $L_2$ norm. On one extreme, for the MNIST dataset, the FGSM is almost entirely accounted for by the adversarially trained neural network, which is expected as it was trained against a PGD attack. Thus, the detector only improved on this by 2.76\%. However, the adversarial training provided little help against the CW attack and the detection system provided most benefit there. For example, in the SVHN case 96.17\% of the robustness was due to the detection system.   

\begin{table}[b]
	\begin{center}
		\caption{Cumulative true and false positive rates for different layer depth over the base network accuracy. }
		
		\begin{tabular}{ | c | c | c | c | }
		 \hline
		    Dataset & 	Layer & \makecell{Cumulative TP increase \\over baseline } & Cumulative FP \\ 
		    \hline
			\multirow{3}{*}{MNIST}&	1 & 76.40\% &  1.28\%  \\ 
			                      &  2 & 82.12\% & 1.58\% \\
			                      & 3 & 84.37\% &  1.96\% \\
			\hline \hline
			\multirow{2}{*}{SVHN}& 1 & 94.06 \% &  6.74\%  \\  
			                     & 2 & 96.17 \% & 10.02\% \\
			\hline
			                    
		\end{tabular}
		\label{Cumul_Table}
	\end{center}
\end{table}

The current threshold used to generate the results gave a FP rate of 1.96\%  on test set data for MINST and  10.02\% for SVHN. However, it is worth highlighting the split between false positives that occur on test set samples that are correctly classified and those which are classified incorrectly. For the current thresholds the number of false positives that occurred on incorrect samples was 0.98\% for MNIST. In other words, half of our false positives in the MNIST case came from data that would have been incorrectly classified. Similarly, for SVHN  3.98\% of the false positives were on incorrectly classified data. Considering a false positive to be only test samples that the neural network classified correctly, then the FP rate is 0.98\% and 6.04\% for MNIST and SVHN. 

We can also examine the TP and FP rates for detectors placed at different depths in the neural network over the baseline accuracy as shown in Table \ref{Cumul_Table} for the CW attack. The results show that if we are including defensive encoders in earlier layers the extra benefit of including additional encoders decreases as fewer adversarial samples are detected. Taken to the extreme with very deep encoders their capability to distinguish adversarial samples is low as by that point if the adversarial sample is misclassified it will be embedded to the point of indistinguishability. 

\subsection{White Box Attacker}

We now turn our attention to a stronger attacker model representing a white box attacker who knows the defensive system in place and hence can mount mount an adaptive attack. Here we restrict our attacker to be constrained to a $L_{\infty}$ norm of 0.3 and 0.05 for MNIST and SVHN respectively.

Such an attacker can consider our detection system as a ensemble of models $i =0,...,k$ and aim to optimise over all of them. Each model is a classifier with encoder-decoders placed in different positions, effectively resembling Fig. \ref{Train_Time_Overview} in the case of $i=1$ but with the auxiliary encoder removed. In the case of $i=0$ then we just have the regular classification network without any encoder-decoders. We combine the model predictions $P_i$, 

\begin{equation}
\label{White_Box_Attacker}
P_{ens} = \sum_{i=0}^{k}P_i(x^*,\theta_i, y)
\end{equation}

and seek to find inputs, $x^*$ which fools the overall ensemble prediction $P_{ens}$. Naively optimising over all the models still leaves the attacker vulnerable to detection. For example, running a PGD attack results in a detection performance of 96.54\% on MNIST. 

The attacker can modify their strategy to specifically counter our proposed defence. A $k$-nn classifier has the obstacle for an attacker that it does not provide gradients to utilise for adversarial sample crafting. Thus, in this attack that we construct the attacker proxies the $k$-nn by not only optimising for misclassification on the network output, but also optimises to push the adversarial embedding as close as possible to the centroid of the target class in embedding space. 
 
To do this we run a PGD attack as it was the best performing $L_{\infty}$ bounded attack. Attack samples which fail at being misclassifed, let alone also being stealthy, are left as they are and do not have further computations performed on them. Successful attacks then have the centroid in latent space of the training data, $X$, for the class they are being misclassifed to, $y_{t}$, computed.   

\begin{figure}[t]
	\centering
	\includegraphics[trim = 0mm 0mm 0mm 0mm, clip, scale=0.30]{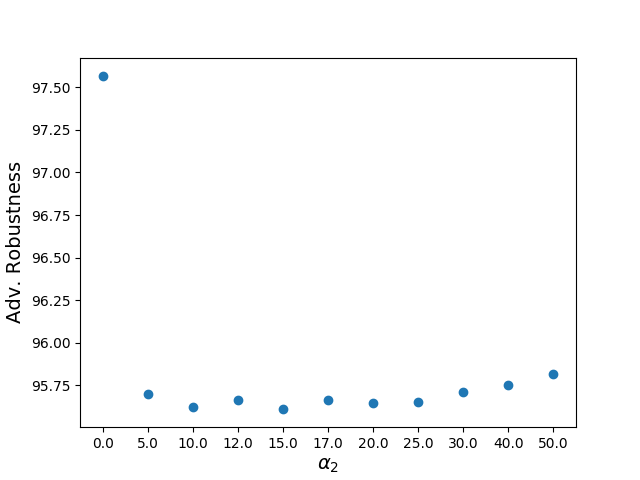}
	\caption{Level of protection offered at different values of $\alpha_2$ for MNIST. Over our search a value of 15 gave the strongest attack on both MNIST and SVHN.}
	\label{L2_LR_Search}	
\end{figure}

We then optimise:

\begin{equation}
\begin{split}
\label{Adaptive_Attacker}
x_{t+1} = x_t + \alpha_{1} \nabla J_{CE}(x^*,y,\theta) \\ 
- \alpha_{2} \sum_{i=1}^{k} \nabla J_{cnt}(x^*,y_{t},X,\theta_i)
\end{split}
\end{equation}

where $J_{cnt}$ is the loss due to the $L_2$ distance in latent space of the target class's centroid to the adversarial sample on a specific encoder $i$. The parameters $\alpha_{1-2}$ weigh the two losses. This type of attack is only feasible under a attacker model that in addition to the target knowledge also has access to a dataset that approximates the training data distribution. In the strongest case, such as we have considered, the attacker has the entirety of the training set. Additionally, success of the attacker under the formulation described depends on how strongly the attacker chooses to optimise for the $J_{cnt}$ loss compared to the objective of successful misclassification. We thus ran a hyper-parameter search over $\alpha_2$ while keeping $\alpha_1$ to 0.01 and the results are shown in Figure \ref{L2_LR_Search} for MNIST.

As we can see, there is a inherent trade off between making the attacks stealthy and yet still achieve the original objective of fooling the classifier. Even under this stronger attack objective our defence is able to detect a high proportion of attacks as shown in Tables \ref{GreyBox}.

\section{Conclusion}

This paper presents Deep Latent Defence, an approach to investigate new ways of defending against adversarial examples. The contribution of this work is to examine how adversarial training can be enhanced in a robust manner with a detection system relying on model confidence. Even under an adaptive $L_{\infty}$ bounded attacker specifically designed to counter our defence we achieved ROC AUC scores of 0.982 and 0.811 for MNIST and SVHN. Under a specific detection threshold this corresponds to a robustness of 95.62\% with a 1.97\% FP rate for MNIST. Equivalent results for SVHN are a robustness of 51.32\% and FP rate of 10.02\%. A substantial number of false positives occur on data that the neural network would have classified incorrectly. Hence, the degradation to the neural network classification performance on normal data is smaller than the raw FP rate, corresponding to 0.98\% and 6.04\% for MNIST and SVHN.

\section*{Acknowledgment}

We would like to thank NVIDIA for their generous donation of a GPU in support of this work. 

\bibliographystyle{unsrt}
\bibliography{main.bib}

\end{document}